\pdfoutput=1
\documentclass{article}
\usepackage{spconf,amsmath,graphicx,algorithm,algorithmic,bm,booktabs,hyperref,multirow}
\usepackage[T1]{fontenc}

\title{Optimal ANN-SNN Conversion with Group Neurons}
\name{Liuzhenghao Lv$^{1}$ \qquad Wei Fang$^{2,3}$ \qquad Li Yuan$^{1,3}$ \qquad Yonghong Tian$^{1,2,3\dagger}$\thanks{This work is supported by the National Natural Science Foundation of China (No.62027804, 62332002, 62202014). \\ $^{\dagger}$ Corresponding author (\href{mailto:yhtian@pku.edu.cn}{yhtian@pku.edu.cn}).}}
\address{$^1$School of Electronic and Computer Engineering, Peking University, China\\
$^2$School of Computer Science, Peking University, China\\
$^3$Peng Cheng Laboratory, China
}
\begin{document}
%\ninept
%
\maketitle
\begin{abstract}
Spiking Neural Networks (SNNs) have emerged as a promising third generation of neural networks, offering unique characteristics such as binary outputs, high sparsity, and biological plausibility. However, the lack of effective learning algorithms remains a challenge for SNNs. For instance, while converting artificial neural networks (ANNs) to SNNs circumvents the need for direct training of SNNs, it encounters issues related to conversion errors and high inference time delays. In order to reduce or even eliminate conversion errors while decreasing inference time-steps, we have introduced a novel type of neuron called Group Neurons (GNs). One GN is composed of multiple Integrate-and-Fire (IF) neurons as members, and its neural dynamics are meticulously designed. Based on GNs, we have optimized the traditional ANN-SNN conversion framework. Specifically, we replace the IF neurons in the SNNs obtained by the traditional conversion framework with GNs. The resulting SNNs, which utilize GNs, are capable of achieving accuracy levels comparable to ANNs even within extremely short inference time-steps. The experiments on CIFAR10, CIFAR100, and ImageNet datasets demonstrate the superiority of the proposed methods in terms of both inference accuracy and latency. Code is available at \url{https://github.com/Lyu6PosHao/ANN2SNN\_GN}.
\end{abstract}

\section{Introduction}
\label{sec:intro}
\begin{figure}[htb]
\begin{minipage}[b]{0.69\linewidth}
  \centering
  \centerline{\includegraphics[width=5.5cm]{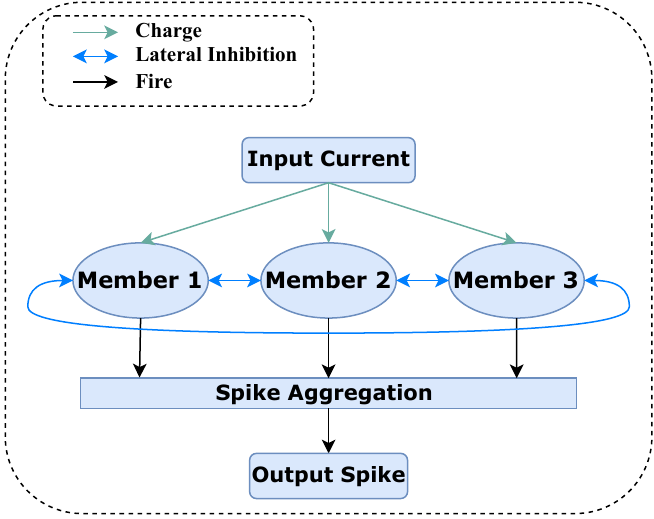}}
%  \vspace{2.0cm}
  \centerline{(a)}\medskip \label{fig:gn}
\end{minipage}
\hfill
\begin{minipage}[b]{0.30\linewidth}
  \centering
  \centerline{\includegraphics[width=2.0cm]{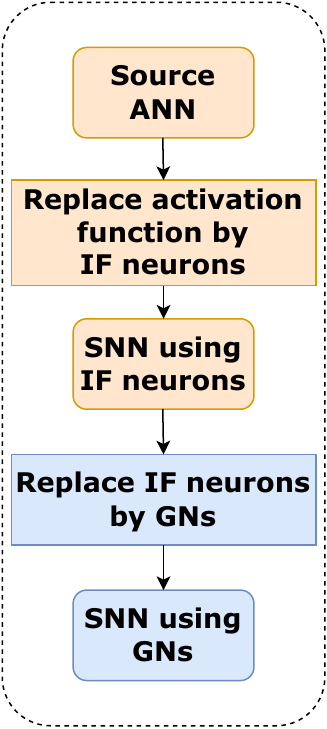}}
%  \vspace{2.0cm}
  \centerline{(b)}\medskip \label{fig:annsnn}
\end{minipage}
\caption{(a) A group neuron composed of three member neurons. (b) Comparison between the traditional ANN-SNN conversion framework and our optimized ANN-SNN conversion framework. In the traditional ANN-SNN conversion framework, the activation functions of the source ANN are typically replaced with IF neurons. Our optimized ANN-SNN conversion framework takes this a step further by replacing IF neurons in the SNN with group neurons.}
\end{figure}
As a new generation of neural networks \cite{maass1997networks}, spiking neural networks (SNNs) have attracted significant attention in the fields of artificial intelligence and computational neuroscience. The key feature of SNNs lies in the use of spiking neurons that closely resemble neurons in the human brain, endowing them with biological plausibility \cite{izhikevich2003simple,mcculloch1943logical}. Similar to real neurons, spiking neurons are activated and generate spikes only when their accumulated membrane potential surpasses a certain threshold. This means that spikes in SNNs are event-driven and highly sparse. Due to the differences between SNNs and artificial neural networks (ANNs), SNNs require specialized hardware in the form of neuromorphic hardware, which has already been researched and developed \cite{debole2019truenorth,davies2018loihi,pei2019towards,xu2022recent}. When deployed on neuromorphic hardware, SNNs exhibit impressive capabilities in processing temporal information, achieving low power consumption, and high efficiency. However, developing efficient learning algorithms for SNNs has been a challenging task.

While the SNN learning algorithms based on direct training often employ surrogate gradients to circumvent the non-differentiability issue of spiking neurons and utilize spatio-temporal back propagation (STBP) for training \cite{cheni2021reducing,wu2018spatio,fang2021deep,feng2022multi}, they typically require significant computational resources and may yield suboptimal model accuracy. In contrast, conversion-based SNN learning methods avoid the direct training of SNNs \cite{cao2015spiking} and tend to achieve higher accuracy compared to directly trained SNNs. However, due to the clipping error, quantization error, and unevenness error introduced during conversion \cite{bu2023optimal}, the accuracy of the converted SNNs is lower than that of the source ANN. This accuracy loss is pronounced, particularly when SNN inference time-steps are not enough. Thus, reducing or even eliminating accuracy loss in ANN-SNN conversion, especially in cases of limited inference time-steps, is crucial for the widespread adoption of SNNs.

Reducing accuracy loss, especially in scenarios with limited SNN inference time-steps, is challenging \cite{rueckauer2017conversion}. Existing ANN-SNN conversion frameworks commonly replace the activation functions in ANNs with Integrate-and-Fire (IF) neurons to obtain SNNs \cite{rueckauer2017conversion, han2020rmp,deng2021optimal,ding2021optimal,ho2021tcl}. This practice relies on the mapping between the ANN activation function values and the firing rates of IF neurons. However, IF neurons have limited expressive capacity, making it difficult to achieve precise mapping within limited time-steps. \cite{li2022efficient} and \cite{wang2022signed} proposed improvements to the IF neurons, but the performances are not good enough.

Alternatively, we propose a novel neuron type called Group Neurons (GNs), as shown in Figure \ref{fig:gn} (a), composed of multiple member neurons with unique neural dynamics that endow it with significantly greater expressive capacity than IF neurons. The firing rates of GNs can accurately map ANN activation values within limited inference time-steps.

Based on GNs, we optimize existing ANN-SNN conversion frameworks. While existing frameworks typically replace ANN activation functions with IF neurons, our method takes it a step further by replacing the IF neurons in the resulting SNN with GNs, as shown in Figure \ref{fig:annsnn} (b). Our experiments demonstrate that SNN using GNs  can achieve almost the same accuracy as that of the original ANN, even in extremely limited time-steps (e.g., 2 time-steps).

% \section{Related Work}
% \label{sec:format}

% The conversion from ANN to SNN is a highly effective method to achieving high-performance SNNs. The earliest research in this field can be traced back to \cite{cao2015spiking}, where they first proposed the mapping between ReLU activation values and the firing frequency of Integrate-and-Fire (IF) neurons for ANN-SNN conversion. Subsequently, \cite{rueckauer2017conversion} conducted in-depth mathematical analysis of the conversion theory, introducing RobustNorm for weight normalization and the use of "reset-by-subtraction" IF neurons to reduce information loss during conversion. \cite{han2020rmp} further analyzed the conversion algorithm from the perspective of residual membrane potential. \cite{deng2021optimal} mathematically analyzed the layer-wise conversion errors and proposed the use of bias terms. \cite{li2021free} performed layer-wise calibration between ANN and SNN. Trainable thresholds have been adopted to achieve more accurate conversion, such as the rate norm layer \cite{ding2021optimal}, trainable clipping layer \cite{ho2021tcl}, and quantization clip-floor-shift activation function \cite{bu2023optimal}. \cite{li2022efficient} and \cite{wang2022signed} proposed improvements to the IF neurons, equipping them with certain capabilities to address unevenness errors, albeit at the cost of sacrificing some biological plausibility. \cite{hao2023bridging} proposed offset spike calibration, which achieved good results but required multiple rounds of calibration before inference, limiting its practical applicability.

\section{Methods}
\label{sec:pagestyle}
In Section \ref{ssecEx}, we introduce the existing ANN-SNN conversion framework and its drawbacks. In Section \ref{secGN}, we propose a novel type of spiking neuron called Group Neurons (GNs). And in Section \ref{secOpt} we use GNs to optimize the existing ANN-SNN conversion framework.
\subsection{Existing ANN-SNN Conversion Framework}
\label{ssecEx}
% 在这一部分，我们介绍了IF神经元以及现有的ANN-SNN转换框架及其理论基础。并且分析了现有ANN-SNN转化框架的缺陷。
In this section, we introduce IF neurons, the existing ANN-SNN conversion frameworks and their drawbacks.

% Integrate-and-Fire Neuron（简称IF神经元）是最简单的一类脉冲神经元，被广泛采用于ANN-SNN转换方法中，这是因为一种映射关系可以在IF神经元的发射频率和ReLU激活值之间建立。IF神经元的动力学行为可以被描述如下：
The Integrate-and-Fire (IF) Neuron is the simplest type of spiking neuron, which is widely utilized in ANN-SNN conversion methods. This is due to the establishment of a mapping relationship between the firing rates of the IF neurons of SNN and the activation values of ANN. The dynamical behavior of the IF neuron can be described as follows:
\begin{equation}
    p^l(t)=v^l(t-1)+W^lx^l(t)  \label{eq-charge}
\end{equation}
\begin{equation}
    s^l(t)=\text{Heaviside}(p^l(t)-\theta^l) \label{eq-spike}
\end{equation}
\begin{equation}
    v^l(t)=p^l(t)-s^l(t)\theta^l  \label{eq-reset}
\end{equation}

% 在这里，公式\ref{eq-charge}描述了IF神经元的充电过程，$v^l(t-1)$表示第$l$层神经元在$t-1$时刻的电位，$W^l$表示第$l-1$层与第$l$层之间的突触连接权重，$x^l(t)$表示第$l$层神经元在$t$时刻的输入电流，$p^l(t)$表示第$l$层神经元在$t$时刻的脉冲发射前电位。公式\ref{eq-spike}描述了IF神经元的放电过程，$Heaviside(\cdot)$函数在自身输入为负值时输出为0，反之则为1。$\theta^l$为第$l$层神经元的阈值，$s^l(t)$为$l$层神经元的输出脉冲。公式\ref{eq-reset}描述了神经元的电位重置过程，在此我们采取了软重置方式，这是因为该方式能够降低ANN-SNN的转换误差。软重置方式意味着如果神经元发射脉冲，则神经元电位将被减去阈值。
Here, Equation  \ref{eq-charge} describes the charging process of the IF neuron, where $v^l(t-1)$ denotes the membrane potential of the neuron in layer $l$ at time-step $t-1$ and $W^l$ denotes the synaptic connection weights between layer $l-1$ and layer $l$. $x^l(t)$ denotes the input current to the neuron in layer $l$ at time-step $t$, and $p^l(t)$ denotes the membrane potential of the neuron in layer $l$ just before spike firing at time-step $t$. 

Equation  \ref{eq-spike} describes spike firing of the IF neuron, where the $\text{Heaviside}(\cdot)$ function outputs 0 if its input is negative and 1 otherwise. $\theta^l$ denotes the threshold of the neuron in layer $l$, and $s^l(t)$ denotes the output spike of the neuron in layer $l$.

Equation  \ref{eq-reset} describes the process of membrane potential reset in the neuron. We adopt soft reset mechanism, as it helps reduce the conversion error in ANN-SNN conversion \cite{rueckauer2017conversion,han2020rmp}. Soft reset implies that if the neuron fires, the membrane potential will be subtracted by the threshold value.

% 如前文所说，在ANN-SNN转换中，我们关心的是IF神经元的脉冲发射频率（或者平均突触后电位）。我们将公式\ref{eq-charge}和\ref{eq-reset}联立：
As mentioned earlier, in ANN-SNN conversion, we are interested in the firing rate of the IF neuron (or the average postsynaptic potential). We can combine equations \ref{eq-charge} and \ref{eq-reset} as follows:
\begin{equation}
    v^l(t)=v^l(t-1)+W^lx^l(t)-s^l(t)\theta^l \label{eq-combine}
\end{equation}

% 之后，我们将上式沿时间$t=1$到$t=T$累加，并且将等式两边同时除以$T$，可以得到：
By summing up the above equation from $t=1$ to $t=T$ and dividing both sides of the equation by $T$, we obtain:
\begin{equation}
    \frac{{v}^{l}(T)-{v}^{l}(0)}{T}=\frac{{W}^{l}\sum_{t=1}^{T} {x}^{l}(t)}{T} - \frac{\sum_{t=1}^{T}{s}^l(t)\theta^{l}}{T}  \label{eq-sum}
\end{equation}

% 我们认为，如果第$l-1$层神经元发射了脉冲，那么第$l$层神经元所接收到的输入为突触后电位$\theta^{l-1}$，即${x}^{l}(t)={s}^{l-1}(t)\theta^{l-1}$，我们将此代入公式\ref{eq-sum}，并且用$\phi^{l}(T)$替换$\frac{\sum_{i=1}^{T}{s}^l(t)\theta^{l}}{T}$：
We assume that if the neuron in layer $(l-1)$ fires, the input received by the neuron in the layer $l$ is given by the postsynaptic potential $\theta^{l-1}$, that is, ${x}^{l}(t)={s}^{l-1}(t)\theta^{l-1}$. We substitute this into Equation  \ref{eq-sum}, and replace $\frac{\sum_{t=1}^{T}{s}^l(t)\theta^{l}}{T}$ with $\phi^{l}(T)$:
\begin{equation}
     \phi^{l}(T)=W^l\phi^{l-1}(T)+(-\frac{{v}^{l}(T)-{v}^{l}(0)}{T}) \label{eq-simple}
\end{equation}

% 由于$\phi^{l}(T)\geq0$，那么$\phi^{l}(T)$可以与ReLU激活函数的激活值$a^l$做映射，映射存在的误差为$|\frac{{v}^{l}(T)-{v}^{l}(0)}{T}|$。可见，推理时间步越短，误差越大。
Since $\phi^{l}(T)\geq0$, $\phi^{l}(T)$ can be mapped to the activation value $a^l$ of the ReLU activation function, with an associated mapping error of $|\frac{{v}^{l}(T)-{v}^{l}(0)}{T}|$. As evident, the fewer the inference time-steps $T$, the larger the mapping error (conversion error).

% \cite{bu2023optimal}在他们的工作中提出了clip-floor-shift (QCFS)激活函数，它被定义为\ref{eq-qcfs}。训练一个使用QCFS激活函数的ANN而不是使用ReLU激活函数的ANN，并将其转换为使用IF神经元的SNN，能有效降低短推理时间步下的转换误差。我们的工作沿用了这种做法出于其优异的表现，不同点在于，我们会将使用SNN中的IF神经元进一步替换为GN，以得到使用GN的SNN作为最终的转换结果。
The quantization clip-floor-shift (QCFS) activation function was introduced by \cite{bu2023optimal}, which is used to reduce the conversion error and is defined in Equation  \ref{eq-qcfs}. Training an ANN with the QCFS activation function instead of ReLU, and subsequently converting it into an SNN using IF neurons, has proven to be effective in reducing conversion errors during short inference time-steps. We use QCFS as the activation function in ANN due to its outstanding performance. 
\begin{align}
    a^l=f(a^{l-1})=\lambda^l clip ( \frac{1}{L}\lfloor \frac{W^la^{l-1} L}{\lambda^l}+0.5\rfloor,0,1 )  \label{eq-qcfs}
\end{align}

Here, $\lambda^l$ denotes the trainable threshold of ANN outputs of layer $l$, which is mapped to the threshold of IF neurons in SNN layer $l$. And $L$ denotes the quantization level of ANN outputs.
% 然而，即使是使用QCFS作为激活函数，在极短的推理时间步下，ANN-SNN转换误差仍然较大。这是因为通过现有的ANN-SNN转换框架获得的SNN均使用IF神经元，而IF神经元的表达能力不强。举例来说，要想让IF神经元的发射频率对激活值0.125进行准确的映射，就至少需要8个推理时间步。在这8个推理时间步内，IF神经元只在某个时间步发射脉冲而其余时间步均不发射。当推理时间步少于8个时，就必然存在转换误差。

%因此，要想在极短的推理时间步下，实现近乎无损的ANN-SNN转换，就必须设计表达能力更强的脉冲神经元。
However, even with the QCFS activation function, ANN-SNN conversion errors persist under very short inference time-steps due to the use of IF neurons, which have limited expressive capacity. For instance, accurately mapping an activation value of 0.125 to IF neuron firing rates requires at least 8 inference time-steps. In these 8 time-steps, the IF neuron must fire once while staying silent in the rest. Fewer than 8 time-steps lead to inevitable conversion errors.

% However, even when using QCFS as the activation function, ANN-SNN conversion errors remain significant under extremely short inference time-steps. This is caused by that the SNNs obtained through existing ANN-SNN conversion frameworks almost all use Integrate-and-Fire (IF) neurons, which have limited expressive capacity. For instance, to accurately map the activation value of 0.125 to the firing rate of an IF neuron, it requires at least 8 inference time-steps. Within these 8 inference time-steps, the IF neuron needs to fire a spike at one time-step while remaining silent during the others. When the number of inference time-steps is less than 8, conversion errors inevitably occur.

In order to achieve nearly lossless ANN-SNN conversion in extremely short inference time-steps, it is imperative to design spiking neurons with greater expressive capacity. Therefore, we have proposed Group Neurons (GNs) to replace IF neurons.

\subsection{Group Neuron}
\label{secGN}
%我们提出了Group Neuron（简称GN），一个GN由$\tau$个IF神经元作为成员来组成。假设被替换的IF神经元的阈值为$\theta^l$，那么GN中的第$i$号成员神经元的阈值$\theta_i^l=\frac{i\theta^l}{\tau}$。GN作为一个整体，其本身的阈值$\theta_{GN}^l=\theta^l$。每个成员神经元具有相同的起始膜电位。

% GN的神经元动力学如下：
One GN is composed of $\tau$ IF neurons as members. Assuming the threshold of the IF neuron to be replaced is $\theta^l$, the threshold of the $i$-th member neuron within the GN is $\theta_i^l=\frac{i\theta^l}{\tau}$. As a whole, the GN has its own threshold, denoted as $\theta_{GN}^l=\frac{\theta^l}{\tau}$. Each member neuron possesses an identical initial membrane potential.

%每个成员神经元具有
The neural dynamics of the GN are as follows:
\begin{equation}
    p^l(t)=v^l(t-1)+W^lx^l(t)  \label{eq-01}
\end{equation}
\begin{equation}
   s_i^l(t)=Heaviside(p^l(t)-\theta_i^l) \ \ \ (\forall i={1,2,\cdots,\tau})  \label{eq-02}
\end{equation}
\begin{equation}
    v^l(t)=p^l(t)-\theta_{GN}^l \sum_{i=1}^\tau{s_i^l(t)}  \label{eq-03}
\end{equation}
\begin{equation}
    s_{GN}^l(t)=\sum_{i=1}^\tau{s_i^l(t)}  \label{eq-04}
\end{equation}

%在这里，公式\ref{eq-01}是每个成员神经元的充电过程，它们的充电电流都是相等的。公式\ref{eq-02}描述了每个成员神经元的发射过程，根据上文提到的，每个成员的阈值是不同的。我们把公式\ref{eq-03}描述的过程称为成员神经元之间的横向抑制，如果某个成员神经元发射了脉冲，那么它不仅需要对自身电压进行软重置，还要对其他所有成员的电压进行软重置。我们把公式\ref{eq-04}描述的过程称为脉冲汇聚，它将所有成员神经元在当前时刻发射的脉冲进行加和得到$s^l(t)$，作为GN的在当前时刻的输出脉冲。 Figure \ref{fig:gn} (a) 展示了一个\tau=3的GN的完整神经元动力学。
% 公式\ref{eq-03}描述了软重置过程。特殊点在于，如果某个成员神经元发射了脉冲，那么它不仅需要对自身膜电位减去$\theta_{GN}^l$，还要对其他所有成员的膜电位减去$\theta_{GN}^l$。因此，我们形象化地把这一过程称为成员神经元之间的横向抑制。

Here, we denote the membrane potentials before spike firing of all member neurons in layer $l$ as \(p^l(t)\) since they are equivalent. For the same reason, we use \(v^l(t)\) to denote the membrane potentials after spike firing of all member neurons. Equation  \ref{eq-01} represents the charging process of each member neuron, with their charging currents being equal. Equation  \ref{eq-02} describes the firing process of each member neuron $i$, with distinct thresholds as previously mentioned. Equation  \ref{eq-03} describes the soft reset process, with a key distinction being that if a member neuron fires, it not only resets its own membrane potential by subtracting $\theta_{GN}^l$ but also resets the membrane potentials of all other members by subtracting $\theta_{GN}^l$. Therefore, we can also conceptually refer to this process as lateral inhibition among member neurons. The process described in Equation \ref{eq-04} is termed spike aggregation. It aggregates the spikes fired by all member neurons at the current time-step to compute $s_{GN}^l(t)$, which serves as the output of the GN at the current time-step.

\begin{figure}[t]
  \centering
  \centerline{\includegraphics[width=9.0cm]{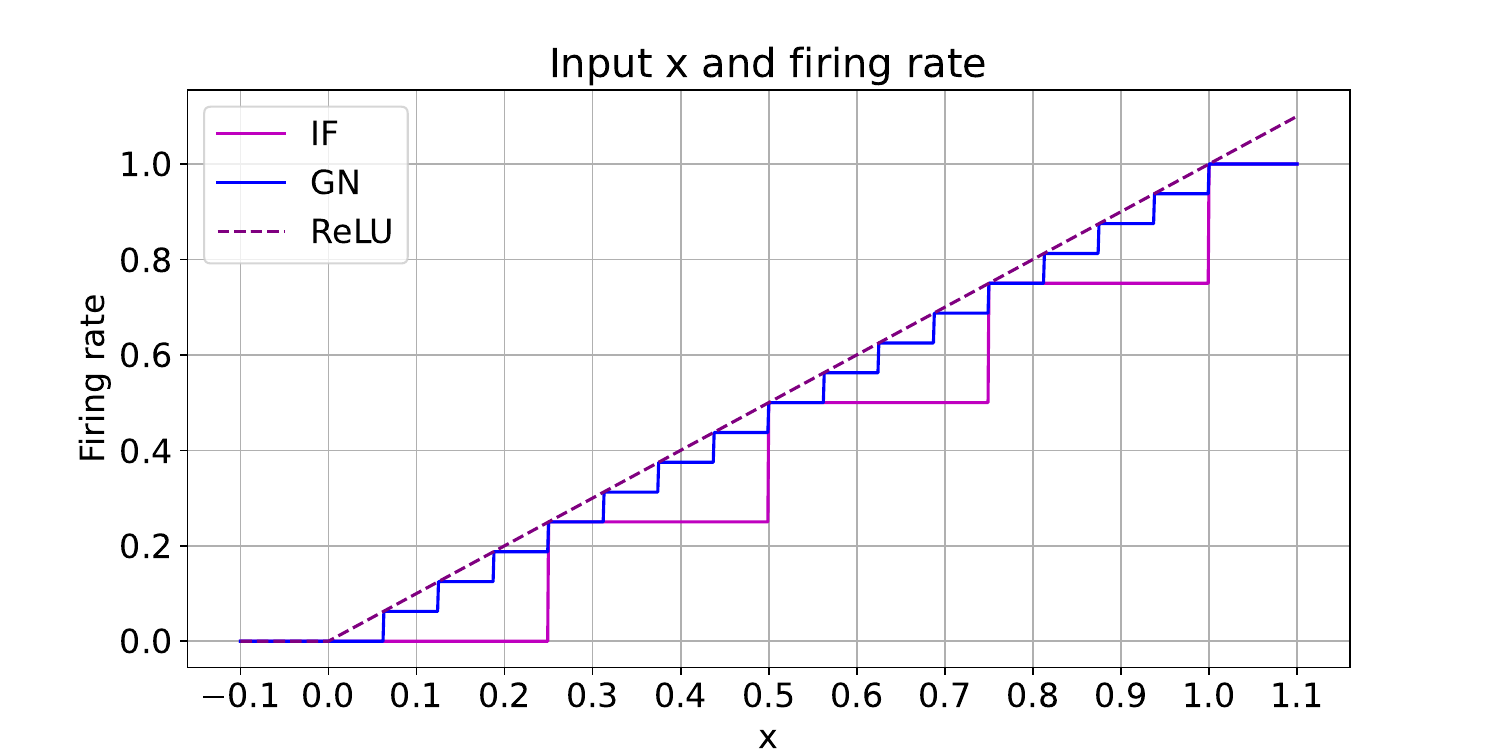}}
  % \centerline{}\medskip

\caption{Comparison between the firing rates of the IF neuron and the GN ($\tau$=4), at the same time-step setting (T=4). }\label{fig:fr}
\end{figure}

Based on equations \ref{eq-01}, \ref{eq-02}, \ref{eq-03}, \ref{eq-04},  following the derivation process in Section \ref{ssecEx}, we can also obtain an equation for the GN, similar to Equation \ref{eq-simple} for the IF neuron:
\begin{equation}
     \phi_{GN}^{l}(T)=W^l\phi_{GN}^{l-1}(T)+(-\frac{{v}^{l}(T)-{v}^{l}(0)}{T}) \label{eq-gngn}
\end{equation}

Here, $\phi_{GN}^{l}(T)=\frac{\sum_{t=1}^{T} {s}_{GN}^l(t)\theta_{GN}^{l}}{T}$. Hence, the $\phi_{GN}^{l}(T)$ of the GN can also be mapped to $a^l$ of the ANN. 
%Compared to IF neurons, GNs exhibit greater expressive capacity, resulting in smaller mapping errors $|\frac{{v}^{l}(T)-{v}^{l}(0)}{T}|$. 

%为了证明这一点，我们绘制了IF神经元与GN神经元的脉冲发射频率随输入电流强度的变化曲线，如图\ref{fig:fr}所示。其中，推理时间步数为4，\tau=4 in the GN，IF神经元的阈值为1。可以观察到，虽然GN和IF神经元的曲线都呈阶梯状，但相较于IF神经元，GN的曲线精细且更加贴近ANN中的激活函数的曲线。这说明，在相同的推理时间步下，GN能达到更小的映射误差，从而降低ANN-SNN转换的精度损失。根据理论计算，
GNs exhibit greater expressive capacity than IF neurons, resulting in smaller mapping errors. We plot the curves of the firing rate as a function of the input current $x$ for the IF neuron and the GN, as shown in Figure \ref{fig:fr}. Here, the number of inference time steps is set to 4, \(\tau\) is set to 4 in the GN , and the threshold of the IF neuron is 1.0. It can be observed that although both GN and IF neuron curves exhibit a step-like pattern, the GN curve is more detailed and closely resembles the activation function curve in ANN when compared to the IF neuron. By derivation, the width of one step of the IF neuron is $\frac{\theta^{l}}{T}$, while that of the GN is $\frac{\theta_{GN}^{l}}{T}$ where $\theta_{GN}^{l}=\frac{\theta^{l}}{\tau}$.
\subsection{Optimal ANN-SNN Conversion Framework}
\label{secOpt}
% 我们优化了现有的ANN-SNN转换框架。以往的做法通常是将源ANN中的激活函数替换为IF神经元，从而得到SNN作为转换的结果。而我们在此基础上更进一步，将该SNN中的IF神经元全部替换为GNs，这个使用GNs的SNN作为转换的最终结果。 

Based on GNs, we have optimized the existing ANN-SNN conversion framework as shown in Figure \ref{fig:annsnn}. The conventional method typically involves replacing the activation functions in the source ANN with IF neurons to obtain an SNN as the conversion result. However, we have taken it a step further by replacing all the IF neurons in this SNN with GNs as shown in Algorithm \ref{algo:replace}, resulting in an SNN that utilizes GNs as the final conversion result.

\begin{algorithm}[htb]
	\caption{\ninept Algorithm for replacing original IF neurons in SNN with GNs}
\ninept
	\label{algo:replace}
    \textbf{Input}: SNN model $M_{\text{SNN}}(\bm{x};\bm{W})$ using IF neurons; Hyper parameter in GNs $\tau$.\\
    \textbf{Output}: $M_{\text{SNN}}(\bm{x};\bm{{W}})\ \text{using GNs}$
	\begin{algorithmic}[1]
 \FOR{$l = 1$ to $M_{\text{SNN}}.\text{layers}$}
        \STATE $M_{\text{SNN}}.\theta_{GN}^l\leftarrow M_{\text{SNN}}.\theta^{l}/\tau$
            \FOR{$i = 1$ to $\tau$}
            \STATE $M_{\text{SNN}}.\theta_{i}^l\leftarrow M_{\text{SNN}}.\theta^{l}*i/\tau$
            \ENDFOR
\ENDFOR
    \RETURN $M_{\text{SNN}}$
    \end{algorithmic}  
    
\end{algorithm}

\section{Experiments}
\label{sec:typestyle}

\subsection{Effectiveness of GN}

% 我们训练一个ResNet-20 on CIFAR100，分别将其转换为使用IF神经元的SNN和使用GNs的SNN。我们分别计算两个SNN与源ANN最后一层的输出值的mean squared error (MSE)，以此衡量转换误差的大小. As shown in fig, 使用IF神经元的SNN的带来的MSE较大，然而使用GNs的SNN带来的MSE要小得多。这说明将IF神经元替换为GNs后，能显著降低转换误差，而且这种效果在推理时间步越短时越显著。
\begin{table}[t]
\centering

\scalebox{0.6}
{
\begin{tabular}{ccccccccc}

\toprule  %添加表格头部粗线
 & T=1& T=2& T=4& T=8& T=16& T=32\\
\midrule  %添加表格中横线
SNN using GN&1.39&	0.64&	0.37&	0.3&	0.28&	0.28 \\
SNN using IF&11.65&	6.36&	3.72	&2.02&	0.92&	0.45 \\

\midrule  %添加表格中横线
ratio& 11.9\%& 10.1\%& 9.9\%& 14.9\%& 30.4\%& 62.2\%&  \\
\bottomrule %添加表格底部粗线
\end{tabular}
}
\caption{MSE of outputs between the last layer of ANN and the last layer of SNN at different time-steps. }\label{tab:neu}%添加标题 设置标签
\end{table}
We train a ResNet-20 on CIFAR-100 and converted it into two types of SNNs: one using IF neurons and the other using GNs. We calculate the mean squared error (MSE) between the final layer outputs of both SNNs and the source ANN to gauge the magnitude of the conversion error. As shown in the Table \ref{tab:neu}, the SNN using IF neurons resulted in a significantly larger MSE than that using GNs. This indicates a substantial reduction in conversion error when replacing IF neurons with GNs. Furthermore, this reduction in conversion error is more pronounced within short inference time-steps, as the ratio is lower at T=1, 2, and 4.

\subsection{Comparison with State-of-the-art Methods}
\begin{table}[htbp]
\centering

\scalebox{0.54}
{
\begin{tabular}{ccccccccc}
\toprule  %添加表格头部粗线
Arch& Method& ANN& T=1& T=2& T=4& T=8& T=16& T=32\\
\toprule  %添加表格头部粗线
\multicolumn{9}{c}{\textbf{\large{CIFAR-10 Dataset}}} \\
\midrule  %添加表格中横线
\multirow{6}{*}{ResNet-18} & QCFS & 95.64\% & 88.84\% & 91.75\% & 93.83\% & 95.04\% & 95.56\% & 95.67\%  \\
 & OPI  & 96.04\% & - & - & - & 75.44\% & 90.43\% & 94.82\%  \\
 & COS  & 95.64\% & 95.25\% & 95.45\% & 95.46\% & 95.66\% & 95.68\% & 95.68\%  \\
 & SRP  & 95.64\% & 94.59\% & 95.06\% & 95.25\% & 95.60\% & 95.55\% & 95.55\%  \\
 & SNNC-AP  & 95.46\% &  - &  - &  - &  -  & -  & 94.78\% \\
 & $\boldsymbol{Ours(\tau=4)}$& 96.48\%& \textbf{96.01\%}& \textbf{96.36\%}& \textbf{96.57\%}& \textbf{96.56\%}& \textbf{96.50\%}& \textbf{96.46\%}  \\
\midrule
\multirow{6}{*}{VGG-16} & QCFS & 95.52\% & 88.41\% & 91.18\% & 93.96\% & 94.95\% & 95.40\% & 95.54\%  \\
 & OPI  & 94.57\% & - & - & - & 90.96\% & 93.38\% & 94.20\%  \\
 & COS  & 95.51\% & 94.90\% & 95.36\% & 95.46\% & 95.51\% & 95.57\% & 95.61\%  \\
 & SRP  & 95.52\% & 93.80\% & 94.47\% & 95.32\% & 95.52\% & 95.44\% & 95.42\%  \\
 & SNNC-AP  & 95.72\% & - & - & - & - & - & 93.71\%  \\
 & $\boldsymbol{Ours(\tau=4)}$& 96.02\%& \textbf{95.63\%}& \textbf{95.84\%}& \textbf{95.94\%}& \textbf{96.00\%}& \textbf{96.02\%}& \textbf{95.97\%}  \\
\midrule  %添加表格中横线
\multicolumn{9}{c}{\textbf{\large{CIFAR-100 Dataset}}} \\
\midrule  %添加表格中横线
\multirow{6}{*}{ResNet-20} & QCFS & 69.94\% & - & 19.96\% & 34.14\% & 55.37\% & 67.33\% & 69.82\%  \\
 & OPI &  70.43\% & - &  - &  -  & 23.09\%  & 52.34\% &  67.18\%  \\
 & COS  & 69.97\% & 59.22\% & 64.21\% & 65.18\% & 67.17\% & \textbf{69.44\%} & \textbf{70.29\%} \\
 & SRP   & 69.94\%  & 46.48\% & 53.96\%  & 59.34\% &  62.94\%  & 64.71\%  & 65.50\% \\
 &  RMP  & 68.72\% &  - &  - &  - &  -  & -  & 27.64\% \\
 & $\boldsymbol{Ours(\tau=4)}$& 68.41\%& \textbf{63.89\%}& \textbf{67.60\%}& \textbf{68.87\%}& \textbf{69.38\%}& {69.15\%}& {69.18\%}  \\
\midrule
\multirow{7}{*}{VGG-16} & QCFS & 76.28\%  & - &  63.79\%  & 69.62\%  & 73.96\%  & 76.24\%  & 77.01\% \\
 & OPI & 76.31\%  & - &  - &  - &  60.49\%  & 70.72\%  & 74.82\%    \\
 & COS  & 76.28\%  & 74.24\%  & 76.03\%  & 76.26\%  & 76.52\%  & 76.77\%  & \textbf{76.96\%}  \\
 & SRP  & 76.28\%  & 71.52\%  & 74.31\%  & 75.42\%  & 76.25\%  & 76.42\%  & 76.45\%  \\
 & SNM  & 74.13\%  &  - &  - &  -  & -  & -  & 71.80\%   \\
 & SNNC-AP  & 77.89\%  &  - &  - &  -  & -  & -  & 73.55\%   \\
 & $\boldsymbol{Ours(\tau=4)}$& 76.43\%& \textbf{75.61\%}& \textbf{76.36\%}& \textbf{76.60\%}& \textbf{76.85\%}& \textbf{76.80\%}& 76.69\%  \\
% \midrule
% ResNet-18 & QCFS & 78.80\% & - & 70.79\% & 75.67\% & \textbf{78.48}\% & \textbf{79.48}\% & \textbf{79.62}\%  \\
% ResNet-18 & SNNC-AP  & 77.89\%  &  - &  - &  -  & -  & -  & 76.32\%   \\
% ResNet-18& $\boldsymbol{Ours(\tau=4)}$& 77.67\%& \textbf{75.92\%}& \textbf{77.77\%}& \textbf{77.95\%}& 78.08\%& 78.02\%& 78.13\%  \\
\midrule  %添加表格中横线
\multicolumn{9}{c}{\textbf{\large{ImageNet Dataset}}} \\
\midrule  %添加表格中横线
\multirow{4}{*}{ResNet-34} & SNNC-AP & 75.36\% & - & - & - & - & - & 64.54\%  \\
 & QCFS & 74.32\% & - & - & - & 35.06\% & 59.35\% & 69.37\%  \\
 & COS  & 74.22\% & 69.11\% & 72.66\% & \textbf{73.81\%} & \textbf{74.17\%} & \textbf{74.14\%} & \textbf{73.93\%} \\
 & SRP & 74.32\% & 57.78\% & 64.32\% & 66.71\% & 67.62\% & 68.02\% & 68.40\%  \\
 & $\boldsymbol{Ours(\tau=6)}$& 74.35\%& \textbf{71.46\%}& \textbf{73.61\%}& {73.73\%}& {73.57\%}& {73.51\%}& {73.46\%}  \\
\bottomrule %添加表格底部粗线
\end{tabular}
}
\caption{Comparison with existing state-of-the-art ANN-SNN conversion methods}\label{tab:compare100}%添加标题 设置标签
\end{table}

We compare our method with other SOTA methods on the CIFAR-10/100 and ImageNet datasets in Table \ref{tab:compare100}, including QCFS \cite{bu2023optimal}, OPI \cite{bu2022optimized}, COS \cite{hao2023bridging}, SRP \cite{hao2023reducing}, RMP \cite{han2020rmp}, and SNNC-AP \cite{li2021free}. On CIFAR-10, for both ResNet-18 and VGG-16, our method outperforms the other methods at the same time-step setting. On CIFAR-100, for ResNet-20, our method outperforms the other methods within limited time-steps (T $\leq$ 8). For VGG-16, we achieve an accuracy almost the same with that of the source ANN with only 2 time-steps (76.36\% vs 76.43\%), which is 0.33\% higher than COS (76.03\%, T=2) and 12.57\% higher than QCFS (63.79\%, T=2). On ImageNet, we achieve an accuracy almost the same with that of the source ANN taking only 2 time-steps (73.61\% vs 74.35\%), while COS taking 4 time-steps (73.81\% vs 74.22\%) and QCFS taking more than 32 time-steps. Note that COS requires 8 additional time-steps for calibrating \cite{hao2023bridging}.
\subsection{Effect of the Number of Member Neurons }
\begin{figure}[t]
  \centering
  \centerline{\includegraphics[width=7.1cm]{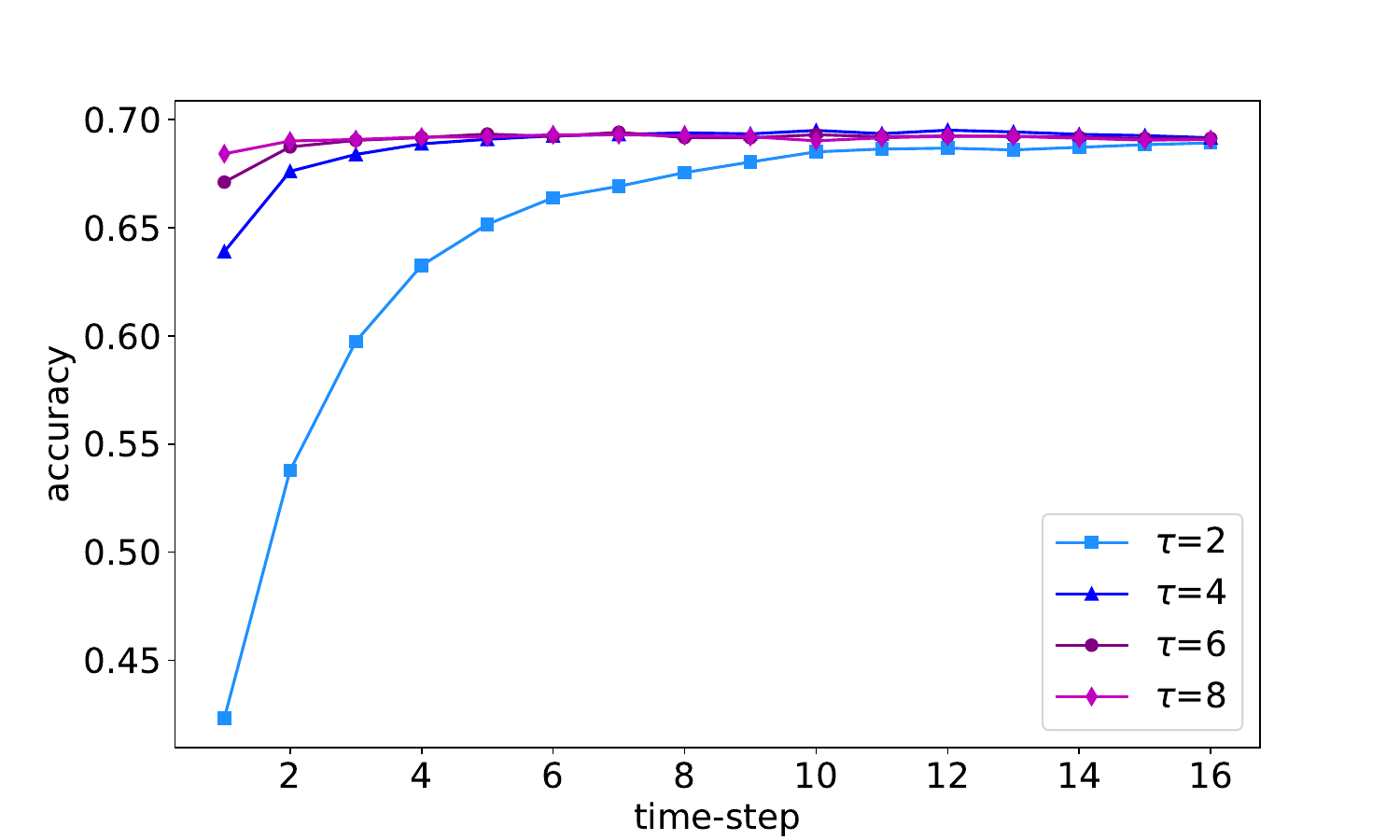}}
  % \centerline{}\medskip
\caption{Effect of different $\tau$. ResNet-20 on CIFAR100.}\label{fig:t}
\end{figure}

Figure \ref{fig:t} illustrates the change in the accuracy for different values of $\tau$. It is evident that increasing $\tau$ significantly improves the accuracy when the time-steps are limited (T $\leq 4$). When there are more time-steps available (T $> 8$), the improvement is no longer as pronounced. 
%Considering that excessively large values of $\tau$ can consume significant GPU memory, a good choice would be $\tau = 4$.

\section{Conclusion}

In this paper, we introduce GNs with enhanced expressive capabilities and optimize the existing ANN-SNN conversion frameworks. Instead of converting the source ANN into an SNN using IF neurons, we convert it into an SNN using GNs. Our method demonstrates excellent performance on the CIFAR-10, CIFAR-100, and ImageNet datasets, achieving low-latency, high-accuracy SNNs. Our method will contribute to the widespread adoption of SNNs.

\newpage

% References should be produced using the bibtex program from suitable
% BiBTeX files (here: strings, refs, manuals). The IEEEbib.bst bibliography
% style file from IEEE produces unsorted bibliography list.
% -------------------------------------------------------------------------
\bibliographystyle{IEEEbib}
\ninept
\bibliography{strings,refs}

\end{document}